\documentclass[letterpaper]{article}
\usepackage{aaai}
\usepackage{times}
\usepackage{helvet}
\usepackage{courier}

\usepackage{amsmath,graphicx}
\usepackage{epstopdf}
\usepackage{multicol,multirow}
\usepackage{subfigure,ctable}
\usepackage{hyperref}
\usepackage{caption}
\usepackage{enumerate} 

\newcommand{\myfigureshrinker}{\vspace{-0.2cm}}

\begin{document}

\title{Authorship Attribution Using a Neural Network Language Model}
\author{Zhenhao Ge, Yufang Sun and Mark J.T. Smith \\
School of Electrical and Computer Engineering, Purdue University\\
465 Northwestern Ave, West Lafayette, Indiana, USA, 47907-2035\\
Emails: \{zge, sun361, mjts\}@purdue.edu, Phone: +1 (317) 457-9348}
\maketitle


\begin{abstract}
\vspace{-0.1cm}
In practice, training language models for individual authors is often expensive because of limited data resources. In such cases, Neural Network Language Models (NNLMs), generally outperform the traditional non-parametric N-gram models. Here we investigate the performance of a feed-forward NNLM on an authorship attribution problem, with moderate author set size and relatively limited data. We also consider how the text topics impact performance. Compared with a well-constructed N-gram baseline method with Kneser-Ney smoothing, the proposed method achieves nearly $2.5\%$ reduction in perplexity and increases author classification accuracy by $3.43\%$ on average, given as few as 5 test sentences. The performance is very competitive with the state of the art in terms of accuracy and demand on test data. The source code, preprocessed datasets, a detailed description of the methodology and results are available at \textrm{\url{https://github.com/zge/authorship-attribution}}.    
\end{abstract}

\vspace{-0.1cm}
\section{Introduction}
\label{sec:intro}

Authorship attribution refers to identifying authors from given texts by their unique textual features. It is challenging since the author's style may vary from time to time by topics, mood and environment. Many methods have been explored to address this problem, such as Latent Dirichlet Allocation for topic modeling \cite{seroussi2011authorship} and Naive Bayes for text classification \cite{coyotl2006authorship}. Regarding language modeling methods, there is mixed advocacy for the conventional N-gram methods \cite{kevselj2003n} and methods using more compact and distributed representations, like Neural Network Language Models (NNLMs), which was claimed to capture semantics better with limited training data \cite{bengio2003neural}. 

Most NNLM toolkits available \cite{mikolov2010recurrent} are designed for recurrent NNLMs which are better for capturing complex and longer text patterns and require more training data. In contrast, the feed-forward NNLM framework we proposed is less computationally expensive and more suitable for language modeling with limited data. It is developed in MATLAB with full network tuning functionalities. 

\begin{figure}[htb]
\myfigureshrinker
\centering
\includegraphics[height=3.2cm, width=7cm]{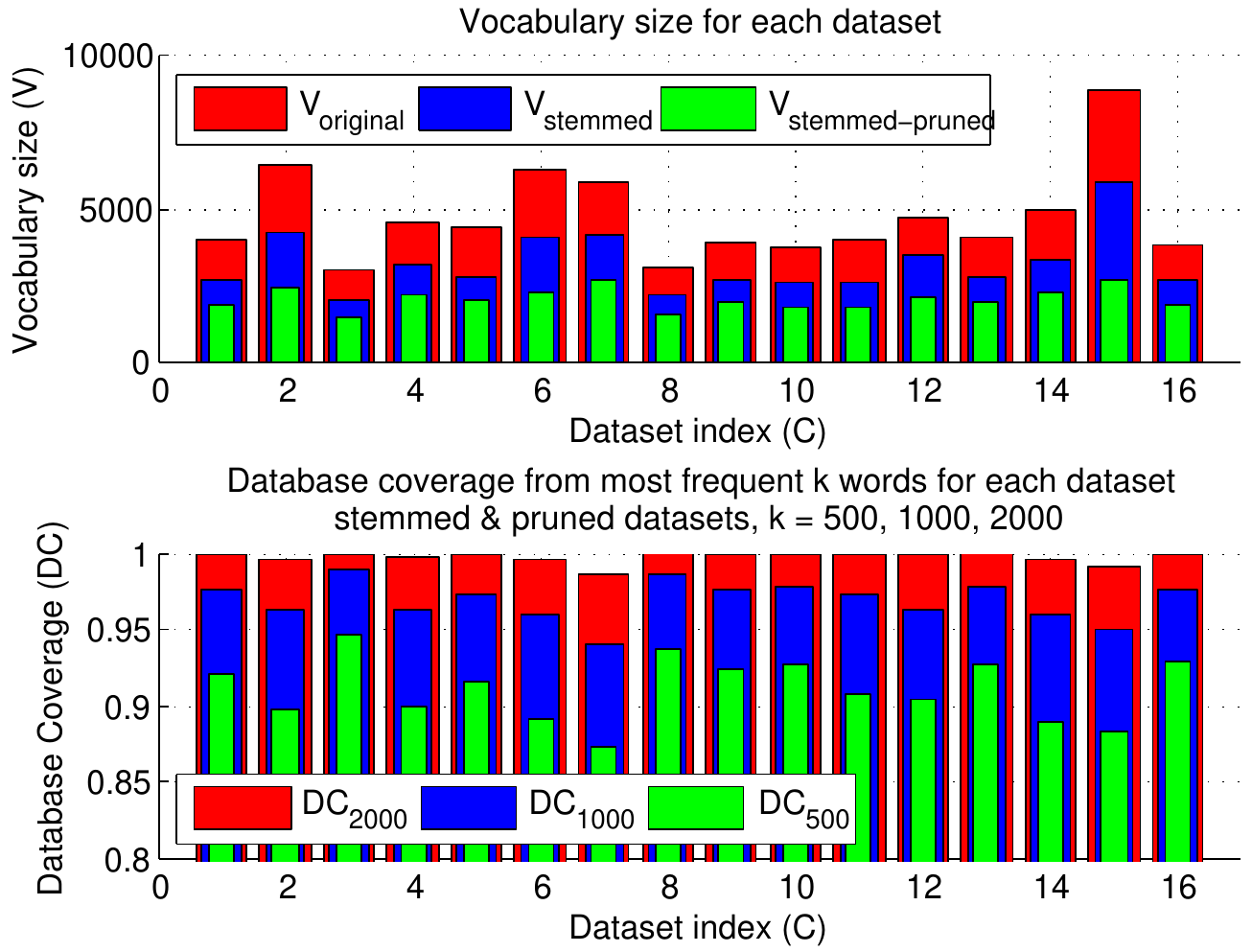}
\caption{Vocabulary size and word coverage in various stages}
\label{fig:data_profile}
\myfigureshrinker
\end{figure}

The database we use is composed of transcripts of 16 video courses taken from Coursera, collected one sentence per line into a text file for each course. To reduce the influence of ``topic'' on author/instructor classification, courses were all selected from science and engineering fields, such as Algorithm, DSP, Data Mining, IT, Machine Learning, NLP, etc. There are 8000+ sentences/course and about 20 words/sentence on average. The vocabulary size of each author varies from $3000$ to $9000$. After stemming with Porter's algorithm and pruning words with frequency less than $1/100,000$, author vocabulary size is reduced to a range from $1800$ to $2700$, with average size around $2000$. Fig. \ref{fig:data_profile} shows the vocabulary size for each course, under various conditions and the database coverage with the most frequent $k$ words $(k=500,1000,2000)$ after stemming and pruning. 

\vspace{-0.1cm}
\section{Neural Network Language Model (NNLM)}

\begin{figure}[tb]
\centering
\includegraphics[height=4.25cm, width=7.5cm]{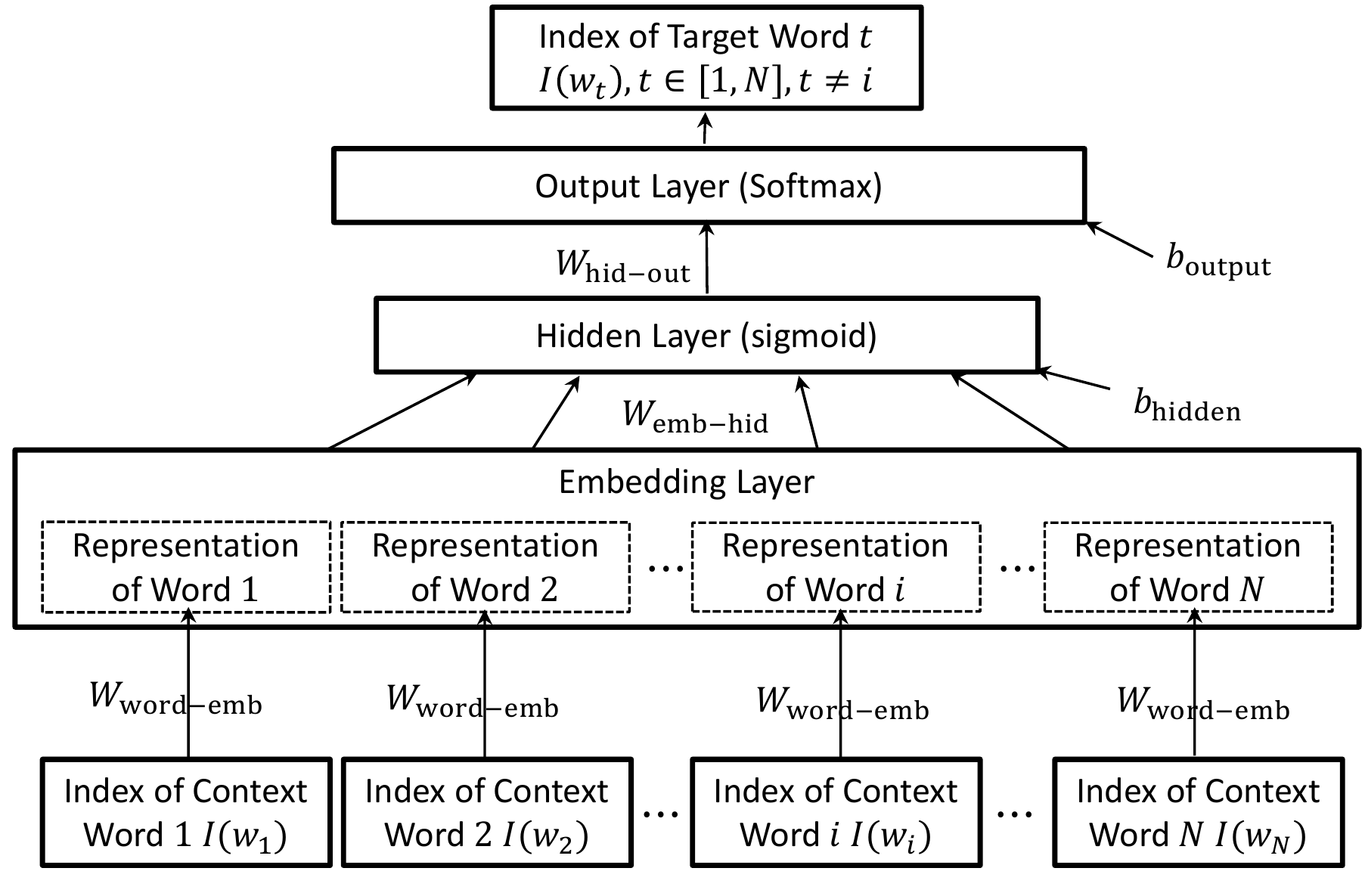}
\caption{A feed-forward NNLM setup ($I$: index, $\mathcal{W}$: word, $N$: number of context words, $W$: weight, $b$: bias)}
\label{fig:nnlm}
\myfigureshrinker
\myfigureshrinker
\end{figure}

Similar to N-gram methods, the NNLM is also used to answer one of the fundamental questions in language modeling: predicting the best target word $\mathcal{W}^{*}$, given a context of $N-1$ words.
%
%
%
The target word is typically the last word within context size $N$. However, it theoretically can be in any position. Fig. \ref{fig:nnlm} demonstrates the structure of the proposed NNLM with multinomial classification cost function:
%
\begin{equation}
\label{eq:costfunction}
C = -\sum_{V} t_{j} \log y_{j} , j \in V , 
\end{equation}
%
where $V$ is the vocabulary size,  $y_j$ and $t_j$ are the final output and the target label. This NNLM setup contains 4 types of layers. The word layer contains $N-1$ input words represented by $V$-dimensional index vectors with $V-1$ ``0''s and one ``1'' positioned in a different location to differentiate it from all other words. Words are then transformed to their distributed representation and concatenated in the embedding layer. Outputs from this layer forward propagate to the hidden sigmoid layer, then softmax layer to predict the probabilities of the possible target words. Weights/biases between layers are initiated randomly and with zeros respectively, and their error derivatives are computed through backward propagation. The network is iteratively updated with parameters, such as learning rate and momentum. 


In implementation, the processed text data for each course are randomly split into training, validation, and test sets with ratio 8:1:1. This segmentation is performed 10 times with different randomization seeds, so the mean/variance of performance of NNLMs can be measured later. We optimized a 4-gram NNLM (predicting the $4^\textrm{th}$ word using the previous 3) with mini-batch training through $10$ to $20$ epochs for each course. The model parameters, such as number of nodes in each layer, learning rate, and momentum are customized for obtaining the best individual models. 

\vspace{-0.1cm}
\section{Classification with Perplexity Measurement}

Denote $\mathcal{W}_{1}^{n}$ as a word sequence $(\mathcal{W}_1,\mathcal{W}_2,\ldots,\mathcal{W}_N)$ and $P(\mathcal{W}_{1}^{n})$ as the probability of $\mathcal{W}_{1}^{n}$ given a LM, perplexity is an intrinsic measurement of the LM fitness defined by:
%
\begin{equation}
\label{eq:ppl1}
PP(\mathcal{W}_{1}^{n}) = P(\mathcal{W}_{1}^{n})^{-\frac{1}{n}}
\end{equation}
%
Using Markov chain theory, $P(\mathcal{W}_{1}^{n})$ can be approximated by the probability of the closest $N$ words $P(\mathcal{W}_{n-N+1}^{n})$, so $PP(\mathcal{W}_{1}^{n})$ can be approximated by
%
\begin{eqnarray}
\label{eq:ppl2}
PP(\mathcal{W}_{n-N+1}^{n}) = ( \prod_{k=1}^{n} P(\mathcal{W}_{k}|\mathcal{W}_{k-N+1}^{k-1}) ) ^ {-1/n}
\end{eqnarray}
%

\begin{figure}[tb]
\centering
\includegraphics[height=3.1cm, width=7cm]{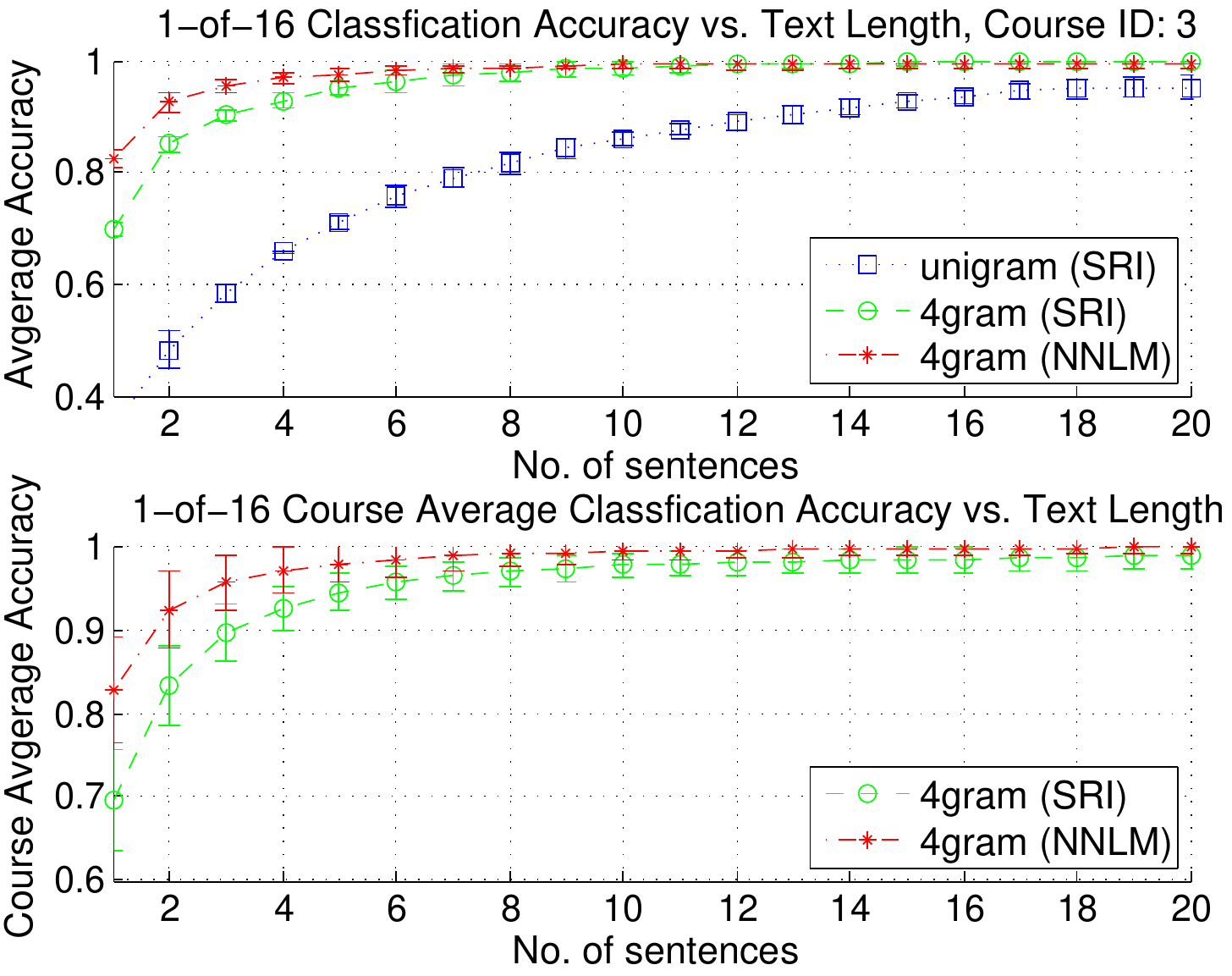}
\caption{Individual and mean accuracies vs. text length in terms of the number of sentences}
\label{fig:acc_sent}
\myfigureshrinker
\end{figure}

\begin{figure}[htb]
\myfigureshrinker
\centering
\includegraphics[height=3.2cm, width=7cm]{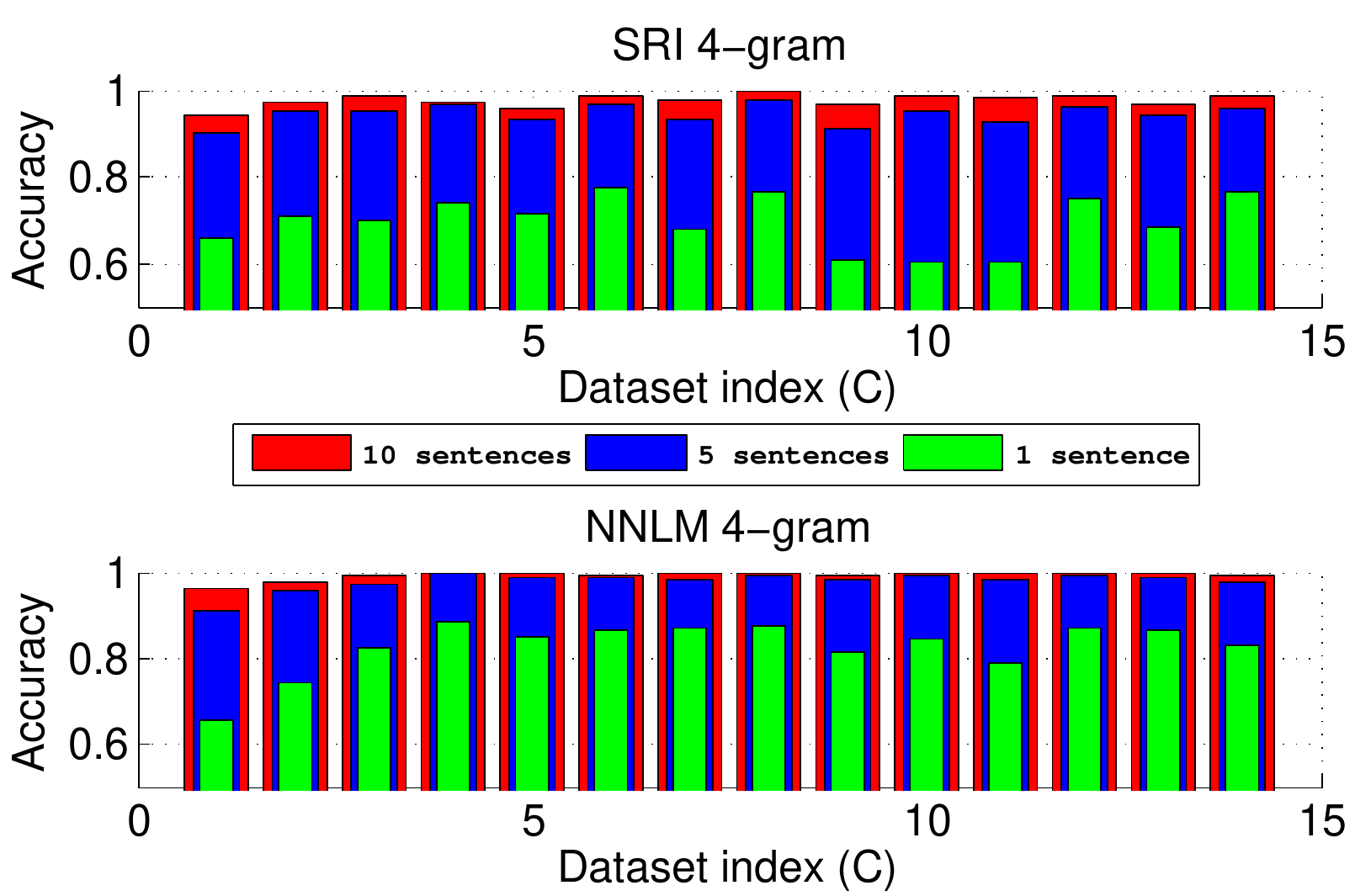}
\caption{Accuracies at 3 stages differed by text length for 14 courses (2 courses from the same instructor are excluded)}. 
\label{fig:acc_stages}
\myfigureshrinker
\myfigureshrinker
\end{figure}

The mean perplexity of applying trained 4-gram NNLMs to their corresponding test sets are $67.3\pm2.4$. This is lower (better) than the traditional N-gram method ($69.0\pm2.4$ with 4-gram SRILM). The classification is performed by finding the author with his/her NNLM that maximizes the accumulative perplexity of the test sentences. By randomly selecting $1$ to $20$ test sentences from the test set, Fig. \ref{fig:acc_sent} shows the 16-way classification accuracy using 3 methods, for one particular course/instructor and for all courses on average. There are 2 courses taught by the same instructor, intentionally added for investigating the topic impact on accuracy. They are excluded when computing the average accuracy in Fig. \ref{fig:acc_sent}. Similarly, the accuracies for courses using two methods with differing text lengths are compared in Fig. \ref{fig:acc_stages}. Both figures show the NNLM method is slightly better than the SRI baselines at the 4-gram level. A classification confusion matrix (not included due to space limits) was also computed to show the similarity between authors. The results show higher confusion on similar courses, which indicates the topic does impact accuracy. The NNLM has higher confusion values than the SRI baseline on the two different courses from the same instructor, so it is more biased toward the author rather than the topic in that sense.

\vspace{-0.1cm}
\section{Conclusion and Future Work}

The NNLM-based work achieves promising results compared with the N-gram baseline. The nearly perfect accuracies given 10+ test sentences are competitive with the state-of-the-art, which achieved 95\%+ accuracy on a similar author size \cite{coyotl2006authorship}, or $80\%+$ with tens of authors and limited training data \cite{seroussi2011authorship}. However, it may also indicate this dataset is not sufficiently challenging, probably due to the  training and test data consistency and the topic distinction. 
In the future, datasets with more authors can be used, for example, taken from collections of books or transcribed speeches. We also plan to integrate a nonlinear function optimization scheme using the conjugate gradient \cite{rasmussen2006gaussian}, which automatically selects the best training parameters and saves time in model customization. To compensate for the relatively small size of the training set, LMs may also be trained with a group of authors and then adapted to the individuals.


\vspace{-0.1cm} 
\small
\bibliography{references}

\begin{thebibliography}{}

\bibitem[\protect\citeauthoryear{Bengio \bgroup et al\mbox.\egroup
  }{2003}]{bengio2003neural}
Bengio, Y.; Ducharme, R.; Vincent, P.; and Janvin, C.
\newblock 2003.
\newblock A neural probabilistic language model.
\newblock {\em The Journal of Machine Learning Research} 3:1137--1155.

\bibitem[\protect\citeauthoryear{Coyotl-Morales \bgroup et al\mbox.\egroup
  }{2006}]{coyotl2006authorship}
Coyotl-Morales, R.~M.; Villase{\~n}or-Pineda, L.; Montes-y G{\'o}mez, M.; and
  Rosso, P.
\newblock 2006.
\newblock Authorship attribution using word sequences.
\newblock In {\em Progress in Pattern Recognition, Image Analysis and
  Applications}. Springer.

\bibitem[\protect\citeauthoryear{Ke{\v{s}}elj \bgroup et al\mbox.\egroup
  }{2003}]{kevselj2003n}
Ke{\v{s}}elj, V.; Peng, F.; Cercone, N.; and Thomas, C.
\newblock 2003.
\newblock N-gram-based author profiles for authorship attribution.
\newblock In {\em PACLING}.

\bibitem[\protect\citeauthoryear{Mikolov \bgroup et al\mbox.\egroup
  }{2010}]{mikolov2010recurrent}
Mikolov, T.; Karafi{\'a}t, M.; Burget, L.; Cernock{\`y}, J.; and Khudanpur, S.
\newblock 2010.
\newblock Recurrent neural network based language model.
\newblock In {\em INTERSPEECH 2010}.

\bibitem[\protect\citeauthoryear{Rasmussen}{2006}]{rasmussen2006gaussian}
Rasmussen, C.~E.
\newblock 2006.
\newblock Gaussian processes for machine learning.

\bibitem[\protect\citeauthoryear{Seroussi, Zukerman, and
  Bohnert}{2011}]{seroussi2011authorship}
Seroussi, Y.; Zukerman, I.; and Bohnert, F.
\newblock 2011.
\newblock Authorship attribution with latent dirichlet allocation.
\newblock In {\em CoNLL}.

\end{thebibliography}
\bibliographystyle{aaai}

\end{document}